\pdfoutput=1

\documentclass[11pt]{article}

\usepackage{acl}

\usepackage{times}
\usepackage{latexsym}

\usepackage[T1]{fontenc}

\usepackage[utf8]{inputenc}

\usepackage{times}
\usepackage{latexsym}

\usepackage{microtype}
\usepackage{hyperref}

\usepackage{caption}
\usepackage{subcaption}

\usepackage{xcolor}
\usepackage{natbib}

\usepackage{latexsym,amsmath,amssymb,amsthm}
\usepackage{cleveref}

\usepackage{mystyle}
\usepackage{symbol}
\usepackage{mlsymbols}

\usepackage{paralist}
\usepackage{sistyle}
\usepackage{tikz}
\usepackage{bbding}
\usepackage[normalem]{ulem}

\newcommand{\bertTiny}{BERT$_{\text{tiny}}$\xspace}
\newcommand{\bertMini}{BERT$_{\text{mini}}$\xspace}
\newcommand{\bertSmall}{BERT$_{\text{small}}$\xspace}
\newcommand{\bertMedium}{BERT$_{\text{medium}}$\xspace}
\newcommand{\bertBase}{BERT$_{\text{base}}$\xspace}
\newcommand{\bertLarge}{BERT$_{\text{large}}$\xspace}
\newcommand{\directprobe}{{\sc DirectProbe}\xspace}

\newcommand{\theTitle}{A Closer Look at How Fine-tuning Changes BERT}

\title{\theTitle}

\author{Yichu Zhou \\
  School of Computing \\
  University of Utah \\
  \texttt{flyaway@cs.utah.edu} \\\And
  Vivek Srikumar \\
  School of Computing \\
  University of Utah \\
  \texttt{svivek@cs.utah.edu} \\}

\begin{document}
\maketitle

\begin{abstract}

    Given the prevalence of pre-trained contextualized
    representations in today's NLP, there have been many
    efforts to understand what information they contain, and why they 
    seem to be universally successful.  The most common approach to use these
    representations involves fine-tuning them for an end task.
    Yet, how fine-tuning changes the
    underlying embedding space is less studied. In this work, we
    study the English BERT family and use two probing
    techniques to analyze how fine-tuning changes the space.
    We hypothesize that
    fine-tuning affects classification performance by
    increasing the distances between examples associated with
    different labels. We confirm this hypothesis with carefully
    designed experiments on five different NLP tasks. Via these experiments,
    we also discover an exception to the prevailing wisdom
    that ``fine-tuning always improves performance''.
    Finally, by
    comparing the representations before and after
    fine-tuning, we discover that fine-tuning does not
    introduce arbitrary changes to representations; instead, it
    adjusts the representations to downstream tasks while
    largely preserving the original spatial structure of the data points.  
    
\end{abstract}


\section{Introduction}\label{sec:introduction}
Pre-trained transformer-based 
language
models~\citep[\eg,][]{devlin-etal-2019-bert}
form the basis of state-of-the-art results across NLP\@.
The relative opacity of these models has prompted 
the development of many probes
to investigate linguistic
regularities captured in them~\citep[\eg,][]{kovaleva-etal-2019-revealing,conneau-etal-2018-cram,jawahar-etal-2019-bert}.

Broadly speaking, there are two ways to use
a pre-trained representation~\cite{peters-etal-2019-tune}: as a fixed feature extractor (where
the pre-trained weights are frozen), or by fine-tuning
it for a task.
The probing literature has largely
focused on the former~\citep[\eg,]
[]{kassner-schutze-2020-negated,perone2018evaluation,
yaghoobzadeh-etal-2019-probing,
krasnowska-kieras-wroblewska-2019-empirical,
wallace-etal-2019-nlp,pruksachatkun-etal-2020-intermediate,aghajanyan-etal-2021-intrinsic}.
 Some previous
work~\citep{merchant-etal-2020-happens,mosbach-etal-2020-interplay,hao-etal-2020-investigating}
does provide
insights about fine-tuning: fine-tuning
changes higher layers more than lower ones and
linguistic information is not lost during fine-tuning.
However, 
relatively less is understood about how the representation
changes \emph{during} the process of fine-tuning and why
fine-tuning invariably seems to improve task performance.


In this work, we investigate the process of fine-tuning
of representations using the English BERT
family~\citep{devlin-etal-2019-bert}. Specifically, we ask:
\begin{enumerate}[nosep]
    \item Does fine-tuning always improve performance?
    \item How does fine-tuning alter the representation to
        adjust for downstream tasks?
    \item How does fine-tuning change the
        geometric structure of different layers?
\end{enumerate}

We apply two probing techniques---classifier-based
probing~\citep{kim-etal-2019-probing,DBLP:conf/iclr/TenneyXCWPMKDBD19}
and \directprobe~\citep{zhou-srikumar-2021-directprobe}---on variants of
BERT representations
that are fine-tuned on five 
tasks: part-of-speech tagging, dependency head
prediction, preposition supersense role \& function
prediction and text classification.
Beyond confirming previous findings about fine-tuning, our analysis 
reveals several new findings, briefly described below.

  First, we find that \textit{fine-tuning introduces a divergence
        between training and test sets}, which is not severe
        enough to hurt generalization in most
        cases. However, we do find one exception where
        fine-tuning hurts the performance; this setting also has the
        largest divergence between training and test set
        after fine-tuning (\Cref{sec:fine-tuned-performance}).
        
        Second, we examine how fine-tuning changes labeled regions of the
        representation space. For a representation where task labels are not
        linearly separable, we find that \textit{fine-tuning adjusts 
        it by 
        grouping points with the
        same label into a small number of  clusters (ideally
        one), thus simplifying the
        underlying representation.} Doing so makes it easier to
        linearly separate labels with fine-tuned
        representations than untuned ones (\Cref{sec:linear}).
        For a representation whose task labels are already
        linearly separable, we find that \textit{fine-tuning 
        pushes the clusters of points
        representing different labels away from each other, thus
        introducing large separating regions between
        labels.} Rather than simply scaling the
        points, clusters move in different directions and
        with different extents (measured by Euclidean distance).
        Overall, these clusters become distant compared to
        the untuned representation.  We conjecture that the
        enlarged region between groups admits a bigger set
        of classifiers that can separate them, leading to
        better generalization (\Cref{sec:findings-strucure}).
        
        We verify our distance hypothesis by investigating the effect of
        fine-tuning across tasks. We observe that
        fine-tuning for related tasks can also provide useful
        signal for the target task by altering the
        distances between clusters representing different
        labels (\Cref{sec:cross-tasks}).
        
        Finally, \textit{fine-tuning does not change the
        higher layers arbitrarily.} This confirms previous
        findings. Additionally, we find that fine-tuning largely
        preserves the relative positions of the label clusters, while
        reconfiguring the space to adjust for downstream
        tasks (\Cref{sec:layers}). Informally, we can say
        that fine-tuning only ``slightly'' changes higher
        layers.

These findings help us understand 
fine-tuning better, and justify why fine-tuned
representations can lead to improvements across many NLP
tasks\footnote{The code and data to replicate our analysis
is available at \url{https://github.com/utahnlp/BERT-fine-tuning-analysis}}.


\section{Preliminaries: Probing Methods}\label{sec:probing}
In this work, we  probe representations in the BERT
family during and after fine-tuning.  First, let
us look at the two supervised probes we will employ: a
classifier-based probe~\citep[\eg,][]{DBLP:conf/iclr/TenneyXCWPMKDBD19,DBLP:journals/corr/abs-2201-10262}
to assess how well a
representation supports classifiers for a task, and
\directprobe~\citep{zhou-srikumar-2021-directprobe} to
analyze the geometry of the representation.

\subsection{Classifiers as Probes}\label{sec:cls}
Trained classifiers are the most commonly used
probes in the literature~\citep[\eg][]{DBLP:conf/emnlp/HewittELM21,whitney2021evaluating,DBLP:journals/corr/abs-2102-12452}.
To understand how well a
representation encodes the labels for a task, a probing
classifier is trained over it, with the embeddings themselves kept frozen when
the classifier is trained.

For all our experiments, we use two-layer neural networks as
our probe classifiers. We use grid-search to choose the
best hyperparameters. Each best classifier is trained five
times with different initializations. We report the average
accuracy and its standard deviation for each classifier. 

The hidden layer sizes are selected from $\{32,
64,128,256\}\times \{32,64,128,256\}$, and the regularizer weight 
from the range $10^{-7}$ to $10^0$. All models use
ReLUs as the activation function for the hidden
layer and are optimized by Adam~\citep{kingma2014adam}. We
set the maximum number of learning iterations to $1000$. We use
\texttt{scikit-learn} v0.22~\cite{scikit-learn} for these experiments.

Classifier probes aim to measure how well a contextualized
representation captures a linguistic property. The
classification performance can help us assess
the effect of fine-tuning.


\subsection{\directprobe: Probing the Geometric Structure}\label{sec:directprobe}
Classifier probes treat the representation as a black box
and only focus on the final task performance; they do not reveal how fine-tuning
changes the underlying
geometry of the space. 
To this end, we use \directprobe~\cite{zhou-srikumar-2021-directprobe}\footnote{We use the \directprobe implementation from
\url{https://github.com/utahnlp/DirectProbe} with default
settings.}, a recently
proposed technique which analyzes embeddings from a
geometric perspective. 
We briefly summarize the technique and refer the reader to the original work for
details.

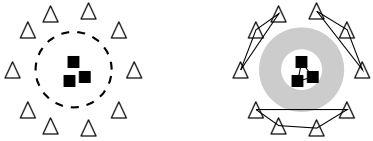
\begin{figure}
  \centering
  \begin{tikzpicture}

      \begin{scope}[shift={(-3,-3)}]
      \node at (4,3.1)  {\tiny $\blacksquare$};
      \node at (3.95,2.85)  {\tiny $\blacksquare$};
      \node at (4.15,2.9)  {\tiny $\blacksquare$};
      \draw [thick, dashed] (4,3) circle [radius=0.5];

      \node at(3.2,3) {$\vartriangle$};
      \node at(3.4,2.471) {$\vartriangle$};
      \node at(3.4,3.529) {$\vartriangle$};
      \node at(3.7,2.258) {$\vartriangle$};
      \node at(3.7,3.742) {$\vartriangle$};
      \node at(4.2,2.225) {$\vartriangle$};
      \node at(4.2,3.775) {$\vartriangle$};
      \node at(4.6,2.471) {$\vartriangle$};
      \node at(4.6,3.529) {$\vartriangle$};
      \node at(4.8,3) {$\vartriangle$};
    \end{scope}

      \begin{scope}[shift={(0,-0.6)}]
      \path [fill=gray!40] (4,0.6) circle [radius=0.56];
      \path [fill=white] (4,0.6) circle [radius=0.27];

      \node (s0) at (4,0.7)  {\tiny $\blacksquare$};
      \node (s1) at (3.95,0.45) {\tiny $\blacksquare$};
      \node (s2) at (4.15,0.5) {\tiny $\blacksquare$};

      \node (c0) at (3.2,0.6) {$\vartriangle$};
      \node (c1) at (3.4,0.071) {$\vartriangle$};
      \node (c2) at (3.4,1.129) {$\vartriangle$};
      \node (c3) at (3.7,-0.142) {$\vartriangle$};
      \node (c4) at (3.7,1.342) {$\vartriangle$};
      \node (c5) at (4.2,-0.175) {$\vartriangle$};
      \node (c6) at (4.2,1.375) {$\vartriangle$};
      \node (c7) at (4.6,0.071) {$\vartriangle$};
      \node (c8) at (4.6,1.129) {$\vartriangle$};
      \node (c9) at (4.8,0.6) {$\vartriangle$};

      \draw[solid] (c0.center) -- (c4.center) -- (c2.center) -- (c0.center);
      \draw[solid] (c1.center) -- (c3.center) -- (c5.center) -- (c7.center) -- (c1.center);
      \draw[solid] (c6.center) -- (c8.center) -- (c9.center) -- (c6.center);
      \draw[solid] (s0.center) -- (s1.center) -- (s2.center) -- (s0.center);

    \end{scope}
    
  \end{tikzpicture}
  \caption{Using the clustering to approximate the set of
    all decision boundaries. The left subfigure is a simple
    binary classification problem with a dashed circular decision
    boundary. The right subfigure is the result of
    \directprobe where the gray area is the region
    that a separator must cross. The connected points
    represent the clusters that \directprobe produces.}\label{fig:lines}
\end{figure}

For a given labeling task,
\directprobe returns a set of clusters such that each
cluster only contains the points with the same label, and
there are no overlaps between the convex hulls of these
clusters.
Any decision boundary must cross the regions
between the clusters that have different labels (see in \Cref{fig:lines}).
Since fine-tuning a contextualized
representation creates different representations for
different tasks, it is reasonable to probe the
representation based on a given task.
These clusters allow us to measure
three properties of interest. 

\noindent\textbf{Number of Clusters}:
The number of clusters indicates the linearity of
the representation for a task. If the number of clusters equals
the number of labels, then examples with the same label are
grouped into one cluster; a simple linear multi-class
classifier will suffice. If, however, there are more clusters than labels,
then at least two clusters of examples with the same label
can not be grouped together (as in \Cref{fig:lines}, right).
This scenario calls for a  non-linear classifier.

\noindent\textbf{Distances between Clusters}:
Distances\footnote{We use
Euclidean distance throughout this work.} between clusters can reveal the
internal structure of a representation. By tracking these
distances during fine-tuning, we can study how the
representation changes. 
To compute these distances, we use the fact that each
cluster represents a convex object. This allows us to use
max-margin separators to compute distances.
We train a linear SVM~\citep{chang2011libsvm} to find the
maximum margin separator and compute its margin.  The
distance between the two clusters is twice the margin.


\noindent\textbf{Spatial Similarity}:
Distances between clusters can also reveal
the spatial similarity of two representations. Intuitively, if two
representations have similar relative distances
between clusters, the representations themselves are
similar to each other for the task at hand. 

We use these distances to construct a \emph{distance vector}
$\mathbf{v}$ for a representation, where each element
$\mathbf{v}_i$ is the distance between the clusters of a
pair of labels. With $n$ labels in a task, the size of
$\mathbf{v}$ is $\frac{n(n-1)}{2}$. This construction works
only when the number of clusters equals the number of labels
(\ie, the dataset is linearly separable under the
representation).  Surprisingly, we find this to be
the case for most representations we studied.  As a measure
of the similarity of two representations for a labeling task,
we compute the Pearson correlation coefficient between their
distance vectors.
Note that this coefficient can also be used to measure the
similarity between two labeled datasets with respect to the
same representation. We exploit this observation to analyze
the divergence between training and test sets for fine-tuned
representations (\Cref{sec:fine-tuned-performance}).



\section{Experimental Setup}\label{sec:exp_setup}
In this section, we describe the representations and
tasks we will encounter in our experiments.

\subsection{Representations}\label{sec:rep}
We investigate several models from the BERT
family~\citep{devlin-etal-2019-bert,turc2019well}. These
models all share the same basic architecture but with different
capacities, \ie, different layers and hidden sizes. \Cref{tb:models} summarizes
the models we investigate in this work\footnote{We ignore
the \bertLarge because, during preliminary experiments, we
found \bertLarge is highly unstable. The variance between
different fine-tuning runs is so large that not comparable
with other BERT models. This is consistent with the
observations from
\citet{DBLP:journals/corr/abs-2006-04884}.}. All of these models are
for English text and uncased.

\begin{table}[]
\small
\centering
\begin{tabular}{@{}lrrrr@{}}
\toprule
            & Layers & \#heads & Dim & \#Param \\ \midrule
\bertTiny   & 2      & 2            & 128       & 4.4M                    \\
\bertMini   & 4      & 4            & 256       & 11.3M                   \\
\bertSmall  & 4      & 8            & 512      & 29.1M                   \\
\bertMedium & 8      & 8            & 512       & 41.7M                   \\
\bertBase   & 12     & 12           & 768      & 110.1M                  \\ \bottomrule
\end{tabular}
\caption{Statistics of five different BERT models.}\label{tb:models}
\end{table}

For tokens that are broken into subwords by the
tokenizer, we average the subword embeddings for the
token representation.  We use the models provided by
HuggingFace v4.2.1~\citep{wolf-etal-2020-transformers}, and
Pytorch v1.6.0~\citep{NEURIPS2019_9015} for our experiments.

\subsection{Tasks}\label{sec:task}


We instantiate our analysis of the BERT models on a diverse
set of five NLP tasks, which covers syntactic and semantic
predictions. Here, we briefly describe the tasks, and refer the reader to the
original sources of the data for further details.\footnote{All the datasets we use in this work are publicly available under
  a creative commons or an open source license. }

\noindent\textbf{Part-of-speech tagging (POS)} predicts the
part-of-speech tag for each  word in a sentence. The task helps us understand if
a representation captures coarse grained syntactic categorization. We use the English
portion of the parallel universal dependencies
treebank~\citep[ud-pud,][]{nivre2016universal}.

\noindent\textbf{Dependency relation (DEP)} predicts the syntactic
dependency relation between two tokens, \ie ($w_{head}$ and
$w_{mod}$). This task can help us understand if, and how
well, a representation can characterize syntactic
relationships between words. This task involves assigning a
category to a \emph{pair} of tokens. We concatenate their
contextualized representations from BERT and treat the
concatenation as the representation of the pair. We use the
same dataset as the POS task for dependencies.


\noindent\textbf{Preposition supersense disambiguation} involves
two categorization tasks of predicting
\emph{preposition's semantic role (\textbf{PS-role})} and
\emph{semantic function (\textbf{PS-fxn})}. These tasks are designed
for disambiguating semantic meanings of prepositions.
Following the previous work~\citep{liu-etal-2019-linguistic},
we only train and evaluate on single-token prepositions from
Streusle v4.2
corpus~\citep{schneider-etal-2018-comprehensive}.

\noindent\textbf{Text classification}, in general, is the task of 
categorizing sentences or documents. 
We use the \textbf{TREC-50} dataset~\citep{li-roth-2002-learning} with 50
semantic labels for sentences. As is the standard practice, we use the
representation of the \texttt{[CLS]} token as the sentence representation. This
task can show how well a representation characterizes a sentence.

\subsection{Fine-tuning Setup}
We fine-tune the models in~\Cref{sec:rep}
on the five tasks from \Cref{sec:task}
separately.\footnote{More detailed settings can be found in
\Cref{sec:fine-tuning-details}} The
fine-tuned models (along with the original models) are then
used to generate contextualized representations. The probing
techniques described in \Cref{sec:probing} are applied to
study both original and fine-tuned representations. 

Our preliminary experiments showed that the commonly used
$3$-$5$ epochs of fine-tuning are insufficient for
the smaller representations, such as \bertTiny, and they
require more epochs. We fine-tuned all the representations
for $10$ epochs except \bertBase, which we fine-tuned for
the usual three epochs. Note that the fine-tuning phase is separate from the 
classifier training phase for probing; for
the probe classifiers, we
train two-layer neural networks (described in \Cref{sec:cls}) from scratch on both original and fine-tuned
representations\footnote{When the fine-tuned representations are
  probed, their weights are frozen. Essentially, after fine-tuning, we treat the
fine-tuned representations as a black-box that produces embeddings for
analysis.}, ensuring a fair comparsion
between them.





\section{Observations and Analysis}\label{sec:findings}

In this section, we will use classifier probes to examine if
fine-tuning always improves classifier performance 
 (\Cref{sec:fine-tuned-performance}). Then
we propose a geometric explanation for \textit{why}
fine-tuning improves classification performance using
\directprobe (\Cref{sec:linear} and \Cref{sec:findings-strucure}).
Next, we will confirm this geometric
explanation by investigating cross-task
fine-tuning (\Cref{sec:cross-tasks}).  Finally, we will analyze how fine-tuning changes the geometry
of different layers of \bertBase (\Cref{sec:layers}).



\subsection{Fine-tuned Performance}\label{sec:fine-tuned-performance}

It is commonly accepted that the fine-tuning improves task
performance. Does this always hold?  \Cref{tb:fine-tuned-performance}
summarizes the relevant observations from our
experiments. \Cref{sec:app-probing-perf} presents the complete fine-tuning results.

\paragraph{Fine-tuning diverges the training and test set.}
In \Cref{tb:fine-tuned-performance}, the last column shows
the spatial similarity between the training and test
set for each representation.  We apply \directprobe on the
training and test set separately. The spatial similarity is
calculated as the Pearson correlation coefficient between
the distance vectors of training and test set (described in
\Cref{sec:probing}).  We observe that after fine-tuning, all
the similarities decrease, implying that the training and
test set diverge as a result of fine-tuning. In
most cases, this divergence is not severe enough to decrease
the performance.

\paragraph{There are exceptions, where fine-tuning hurts
performance.} An interesting observation in \Cref{tb:fine-tuned-performance} is that \bertSmall does not show the
improvements on the PS-fxn task after fine-tuning, which
breaks the well-accepted impression that fine-tuning always
improve the performance. However, only one such exception is
observed across all our experiments (see \Cref{sec:app-probing-perf}).
It is insufficient to draw any concrete conclusions
about why this is happening.
We do  observe that \bertSmall shows the smallest
similarity ($0.44$) between the training and test set after
fine-tuning on PS-fxn task.
We conjecture that controlling the divergence
between the training and test sets can help ensure that fine-tuning
helps. Verifying or refuting this conjecture requires further study.



\begin{table}[]
\centering
\small
\begin{tabular}{@{}llrr@{}}
\toprule
Task                    &          & Acc   & Sim  \\ \midrule
\multirow{2}{*}{POS}     & original & 94.25 & 0.96 \\
                         & tuned    & 94.43 & 0.72 \\ \midrule
\multirow{2}{*}{DEP}     & original & 92.93 & 0.93 \\
                         & tuned    & 94.48 & 0.78 \\ \midrule
\multirow{2}{*}{PS-fxn}  & original & 86.26 & 0.82 \\
                         & tuned    &
                         \underline{\textit{85.08}} &
                         \textbf{0.44} \\ \midrule
\multirow{2}{*}{PS-role} & original & 74.22 & 0.84 \\
                         & tuned    & 74.57 & 0.54 \\ \midrule
\multirow{2}{*}{TREC-50} & original & 81.32 & -  \\
                         & tuned    & 89.60 & - \\ \bottomrule
\end{tabular}\caption{Fine-tuned performances of
\bertSmall based on the last layers. The last
column shows the spatial similarity (described in
\Cref{sec:probing}) between the training and
test set. A complete table of all representations and tasks
can be found in \Cref{sec:app-probing-perf}.
}\label{tb:fine-tuned-performance}
\end{table}

\subsection{Linearity of Representations}\label{sec:linear}

Next, let us examine the geometry of the representations
before and after fine-tuning using \directprobe and
counting the number of clusters. We will focus on the overwhelming majority of
cases where fine-tuning does improve performance.

\paragraph{Smaller representations require more complex
classifiers.} \Cref{tb:linearity} summarizes the results.
For brevity, we only present the results on \bertTiny. The
full results are in \Cref{sec:app-probing-perf}.  We observe
that before fine-tuning, small representations (\ie,
\bertTiny) are non-linear for most tasks. Although a
non-linearity does not imply poor generalization, it
represents a more complex spatial structure, and requires a
more complex classifier.  This suggests that to use small
representations (say, due to limited resources), it would be
advisable to use a non-linear classifier rather than a
simple linear one.

\paragraph{Fine-tuning makes the space simpler.} In \Cref{tb:linearity},
we observe that the number of clusters decreases
after fine-tuning. This tells us that after fine-tuning, the
points associated with different labels are in a simpler
spatial configuration. The same trend holds for TREC-50 (\Cref{tb:TREC-50-linearity}), even when the final
representation is \emph{not} linearly separable.

\begin{table}[h]
\centering
\footnotesize
\begin{tabular}{@{}llrrr@{}}
\toprule
 Task                 &            & \#clusters & is linear & Acc\\ \midrule
    \multirow{2}{*}{POS} & original   & 3936      & N   & $90.76$      \\
                         & tuned      & 20        & N   & $91.67$       \\ \midrule
 \multirow{2}{*}{DEP} & original   & 653       & N    & $86.74$      \\
                      & tuned      & 46        & Y    & $89.04$     \\  \midrule
 \multirow{2}{*}{PS-fxn} & original   & 402       & N  & $74.14$       \\
                      & tuned    & 40        & Y  & $74.40$       \\  \midrule
    \multirow{2}{*}{PS-role}  & original   & 46        & Y &  $58.38$      \\
                           & tuned & 46        & Y  & $60.31$      \\ \midrule
    \multirow{2}{*}{TREC-50} & original & 399 & N & $68.12$ \\
                            & tuned & 51 & N & $84.04$ \\
                           \bottomrule
\end{tabular}\caption{The linearity of the last layer of
    \bertTiny for each task. Other results are in
    \Cref{sec:app-probing-perf}.}\label{tb:linearity}
\end{table}

\begin{table}[]
\centering
\footnotesize
\begin{tabular}{@{}llrrr@{}}
\toprule
Rep                     &          & \#clusters & is linear & Acc   \\ \midrule
\multirow{2}{*}{\bertTiny}   & original & 399        & N         & 68.12 \\
                        & tuned    & 51         & N         & 84.04 \\ \midrule
\multirow{2}{*}{\bertMini}   & original & 127        & N         & 74.12 \\
                        & tuned    & 52         & N         & 88.36 \\ \midrule
\multirow{2}{*}{\bertSmall}  & original & 113        & N         & 81.32 \\
                        & tuned    & 51         & N         & 89.60 \\ \midrule
\multirow{2}{*}{\bertMedium} & original & 110        & N         & 80.68 \\
                        & tuned    & 52         & N         & 89.80 \\ \midrule
\multirow{2}{*}{\bertBase}   & original & 162        & N         & 85.24 \\
                        & tuned    & 51         & N         & 90.36 \\ \bottomrule
\end{tabular}\caption{The linearity of the last layer of
    all models on TREC-50 task. The number of clusters is always more than the
    number of labels (50). }\label{tb:TREC-50-linearity}
\end{table}


\subsection{Spatial Structure of Labels}\label{sec:findings-strucure}


To better understand the changes in spatial structure, we
apply \directprobe to \emph{every} intermediate representation
encountered during fine-tuning. Here, we focus on the
\bertBase. Since all representations we considered are
linearly separable\footnote{In this part, we
exclude the TREC-50 task because it is non-linear even after
fine-tuning. It is difficult to track the minimum distances
between clusters when the clusters are merging during
fine-tuning.}, the number of clusters equals the number
of labels. As a result, each cluster exclusively corresponds to one label. Going ahead, we will use clusters
and labels interchangeably.

\paragraph{Fine-tuning pushes each label far away from each
other.} This confirms the observation of
\Citet{zhou-srikumar-2021-directprobe}, who pointed out that the
fine-tuning pushes each label away from each other.
However, they use the global minimum distance between
clusters to support this argument, which only partially
supports the claim: the distances between some clusters
might increase despite the global minimum distance decreasing.

\begin{figure}[h]
    \centering
    \includegraphics[width=0.45\textwidth]{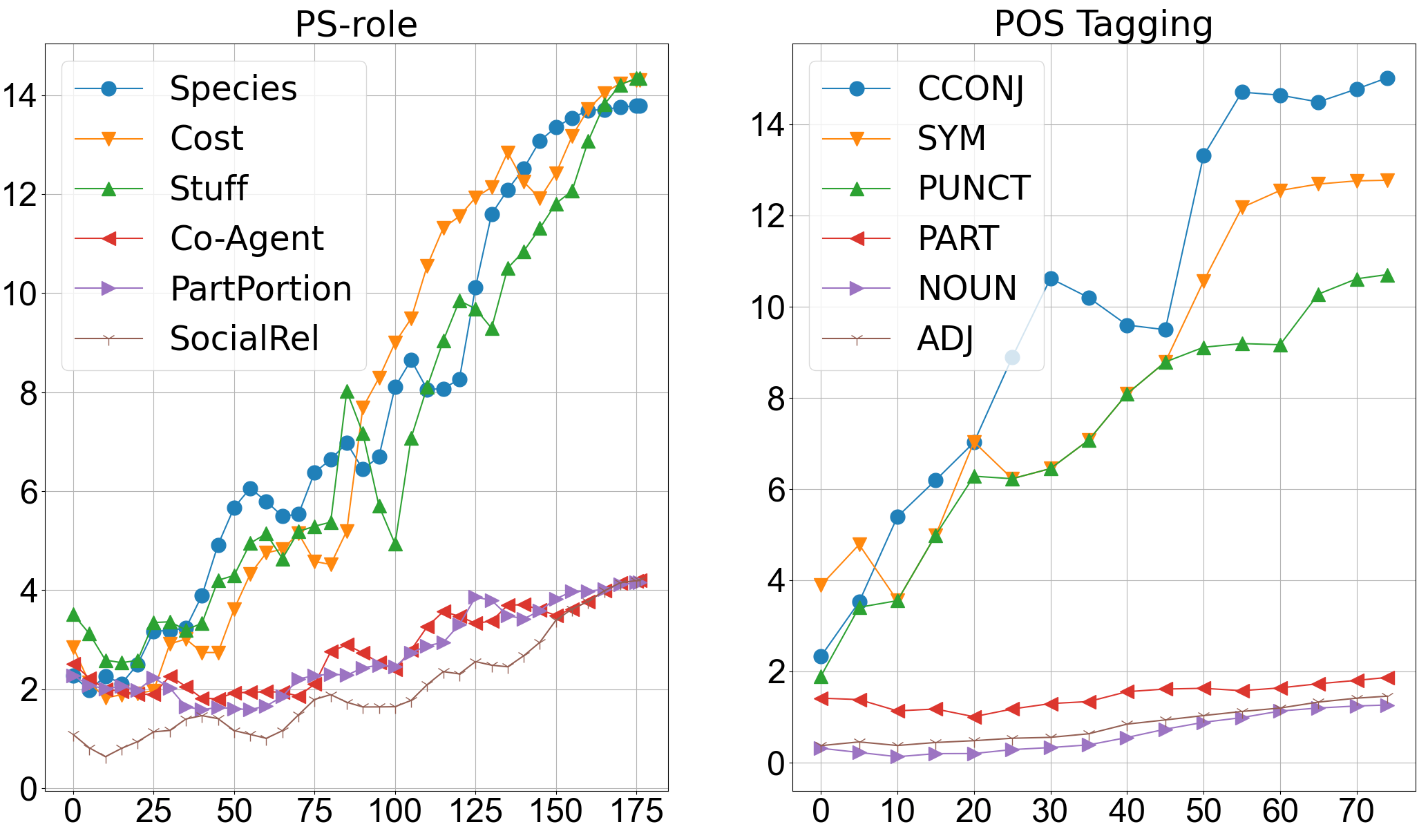}
    \caption{The dynamics of the minimum distances of the
    three labels where the distance increases the most, and
    the three where it increases the least. The horizontal
    axis is the number of fine-tuning updates; the vertical
    axis is chosen label's minimum distance to other labels.
    These results come from the last layer of \bertBase. A
    full plots of four tasks can be found
    in \Cref{sec:app-dynamic-min-dis}.}\label{fig:dynamic_dis}
\end{figure}

We track the minimum distance of each label to all other
labels during fine-tuning. We find that all the minimum
distances are increasing. \Cref{fig:dynamic_dis} shows
how these distances change in the last layer of \bertBase
for the PS-role and POS tagging tasks. \Cref{sec:app-dynamic-min-dis}
includes the plots for all tasks.  For clarity, we only show
the three labels where the distance increases the most, and
the three where it increases the least.  We also observe
that although the trend is increasing, the minimum distance
associated with a label may decrease during the course of fine-tuning,
\eg, the label \textsc{Stuff} in PS-role task, suggesting
a potential instability of fine-tuning.

\begin{figure}
    \centering
    \includegraphics[width=0.45\textwidth]{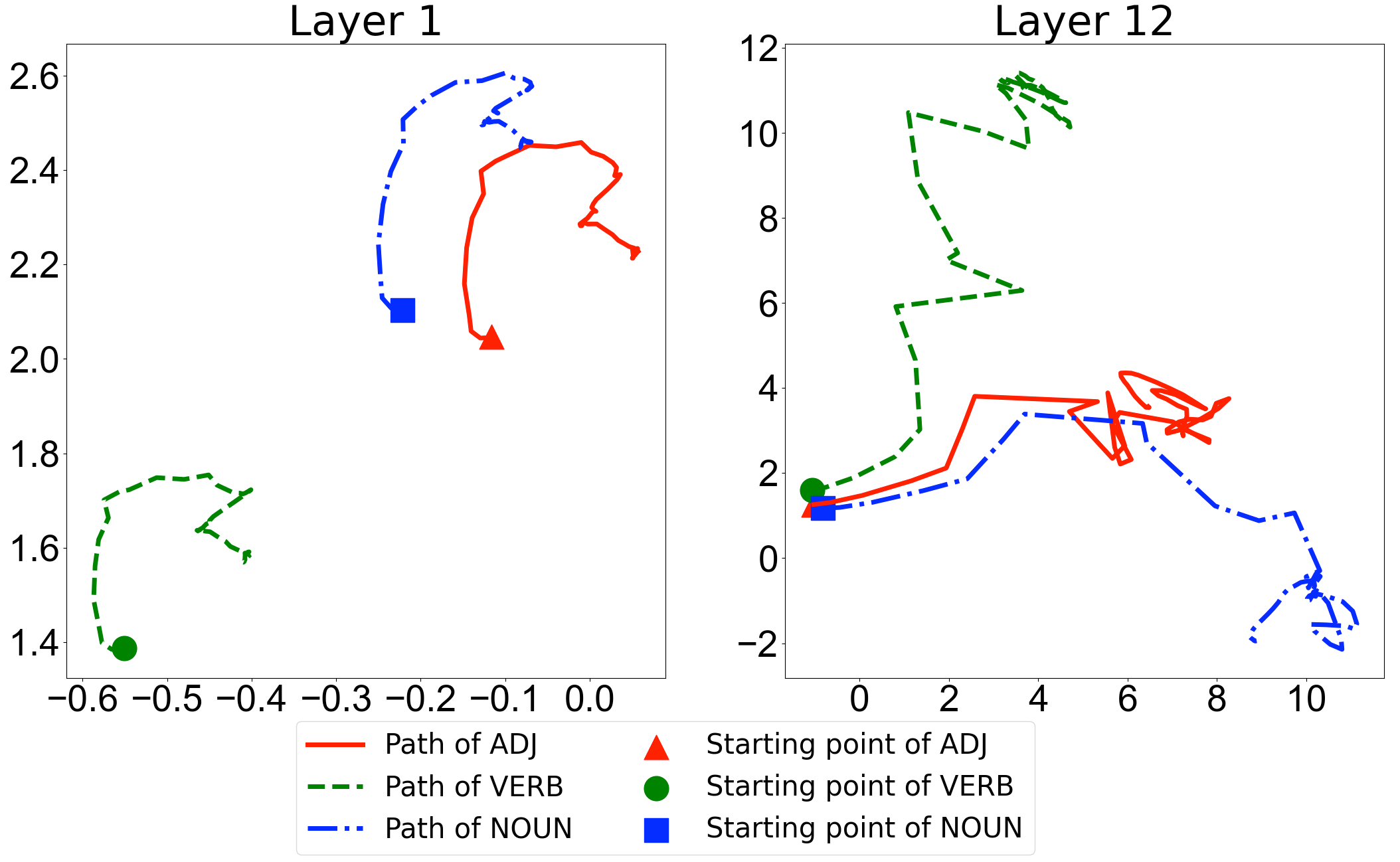}
    \caption{The PCA projection of three closest labels in
    POS tagging task based on the first (left) and last
    (right) layer of \bertBase.  These lines show the paths
    of the centroids of each label cluster during the
    fine-tuning.  The markers indicate the starting
    points. This figure is best seen in color.}\label{fig:pca_paths}
\end{figure}

To further see how labels move during the fine-tuning,
we track the centroids of each cluster. We select three closest labels
from the POS tagging task and track the paths of the
centroids of each label cluster in the last layer of
\bertBase during the fine-tuning. 
\Cref{fig:pca_paths} (right) shows the 2D PCA projection
of these paths. We observe that before fine-tuning, the centroids of
all these three labels are close to each other. As
fine-tuning proceeds, the centroids move around in different
directions, away from each other.

We conclude that fine-tuning enlarges the gaps between label
clusters and admits more classifiers consistent
with the labels, allowing for better generalization.
Note that neither the loss nor the optimizer \emph{explicitly}
mandates this change. Indeed,
since the labels were originally linearly separable, the
learner need not adjust the representation at all.


\subsection{Cross-task Fine-tuning}\label{sec:cross-tasks}
In~\Cref{sec:findings-strucure}, we hypothesized that fine-tuning
improves the performance because it enlarges the gaps
between label clusters. A natural inference of this
hypothesis is that the process may shrink the gaps between
labels of an unrelated task, and its performance can decrease. In this subsection,
we investigate how fine-tuning
for one task affects another.

We fine-tune the \bertBase on PS-role and POS tagging tasks
separately and use the fine-tuned models to generate
contextualized representations for the PS-fxn task. Our
choice of tasks in this experimental design is motivated by
the observation that PS-role and
PS-fxn are similar tasks that seek to predict
supersense tags for prepositions. On the other hand, POS
tagging can adversely affect the PS-fxn task because POS
tagging requires all the prepositions to be grouped together
(label \textsc{ADP}) while PS-fxn requires different prepositions to
be far away from each other. We apply \directprobe on
both representations to analyze the geometric
changes\footnote{The PS-fxn task is still linearly
separable even after fine-tuning on PS-role or POS tagging
tasks.} with respect to PS-fxn. 


\paragraph{The effects of cross-task fine-tuning depends on
how close two tasks are.}
The third and fourth columns of \Cref{tb:cross-tasks} indicate
the number of labels whose minimum distance is increased or
decreased after fine-tuning. The second column from the
right shows
the average distance change over all labels, \eg fine-tuning
on POS results in the minimum distances of the PS-fxn
labels decreasing by $1.68$ on average. We observe that
fine-tuning on the same dataset (PS-fxn)
increases the distances between labels (second row), which is
consistent with observations from \Cref{sec:findings-strucure};
fine-tuning on a similar task also increases the distances
between clusters (third row) but to a lesser extent.
However, fine-tuning on a ``opposing'' task decreases the distances
between clusters (last row). These observations suggest that
cross-task fine-tuning could add or remove information from the
representation, depending on how close the
source and target task are.

\paragraph{Small distances between label clusters indicate a
poor performance.} Based on our conclusion in \Cref{sec:findings-strucure}
that a larger gap between
labels leads to better generalization, we expect that
the performance of PS-fxn after fine-tuning on PS-role would
be higher than the performance after fine-tuning on POS
tagging. To verify this, we train two-layer neural networks
on PS-fxn task using the representations that are fine-tuned on
PS-role and POS tagging tasks. Importantly, we do not further
fine-tune the representations for PS-fxn. The last column of
\Cref{tb:cross-tasks} shows the results. Fine-tuning on
PS-fxn
enlarges gaps between \textit{all} PS-fxn labels, which
justifies the highest performance; fine-tuning on PS-role
enlarges gaps between \textit{some} labels in PS-fxn,
leading to a slight improvement;
fine-tuning on POS tags shrinks the gaps between
\textit{all} labels in PS-fxn, leading to a decrease in
performance.  

In summary, based on the results of \Cref{sec:linear},
\Cref{sec:findings-strucure} and \Cref{sec:cross-tasks}, we
conclude that fine-tuning injects or removes task-related information
from representations by adjusting the distances between
label clusters \emph{even if} the original representation is
linearly separable (i.e., when there is no need to change the representation). When the original
representation does not support a linear classifier,
fine-tuning tries to group points with the same label into a
small number of clusters, ideally one cluster.

\begin{table}[]
\centering
\small
\begin{tabular}{@{}llrrrr@{}}
\toprule
    fine-tuning & probing & \#inc & \#dec & average inc &
    Acc    \\ \midrule
    -           & PS-fxn &  -     & -     & -           & 87.75  \\
PS-fxn  & PS-fxn   &  40    & 0     & 5.29        & 89.58 \\
PS-role   & PS-fxn  &  27    & 13    & 1.02        & 88.53  \\
POS & PS-fxn &  0     & 40    & -1.68       & 83.24 \\ \bottomrule
\end{tabular}\caption{Classification performances for PS-fxn
    task using the last layer of \bertBase when fine-tuning
    on different tasks. First row indicates the untuned
    version. The third and forth
    column indicate the number of labels whose minimum
    distance is increased or decreased after fine-tuning.
    The second last column (average inc) shows the average change
    of the minimum distance over all the labels. The last
    column indicates the probing accuracy.}\label{tb:cross-tasks}
\end{table}


\subsection{Layer Behavior}\label{sec:layers}
Previous
work~\citep{merchant-etal-2020-happens,mosbach-etal-2020-interplay}
showed that during fine-tuning, lower layers changed little
compared to higher layers. In the following experiments, we 
confirm their findings and further show that:
\begin{inparaenum}[(i)]
    \item fine-tuning does not change the representation
  arbitrarily, even for higher layers;
    \item an analysis of the changes of
        different layers by a visual comparison between
        lower and higher layers.
\end{inparaenum}
Here, we focus on the POS tagging task with
\bertBase.
Our conclusions extend to other tasks,
whose results are in \Cref{sec:app-pca-movements}.

%
%

\paragraph{Higher layers do not change arbitrarily.}
Although previous work~\citep{mosbach-etal-2020-interplay}
shows that higher layers change more than the lower layers,
we find that higher layers still remain close to the
original representations.  To study the dynamics of
fine-tuning, we compare each layer during fine-tuning to its
corresponding original pre-trained one. The spatial
similarity between two representations is calculated as the
Pearson correlation coefficient of their distance vectors as
described in~\Cref{sec:probing}.  Intuitively, a classifier
learns a decision boundary that traverses the region between
clusters, which makes the distances between clusters more
relevant to our analysis (as opposed to the spatial structure of points within each cluster).

\begin{figure}[h]
    \centering
    \includegraphics[width=0.45\textwidth]{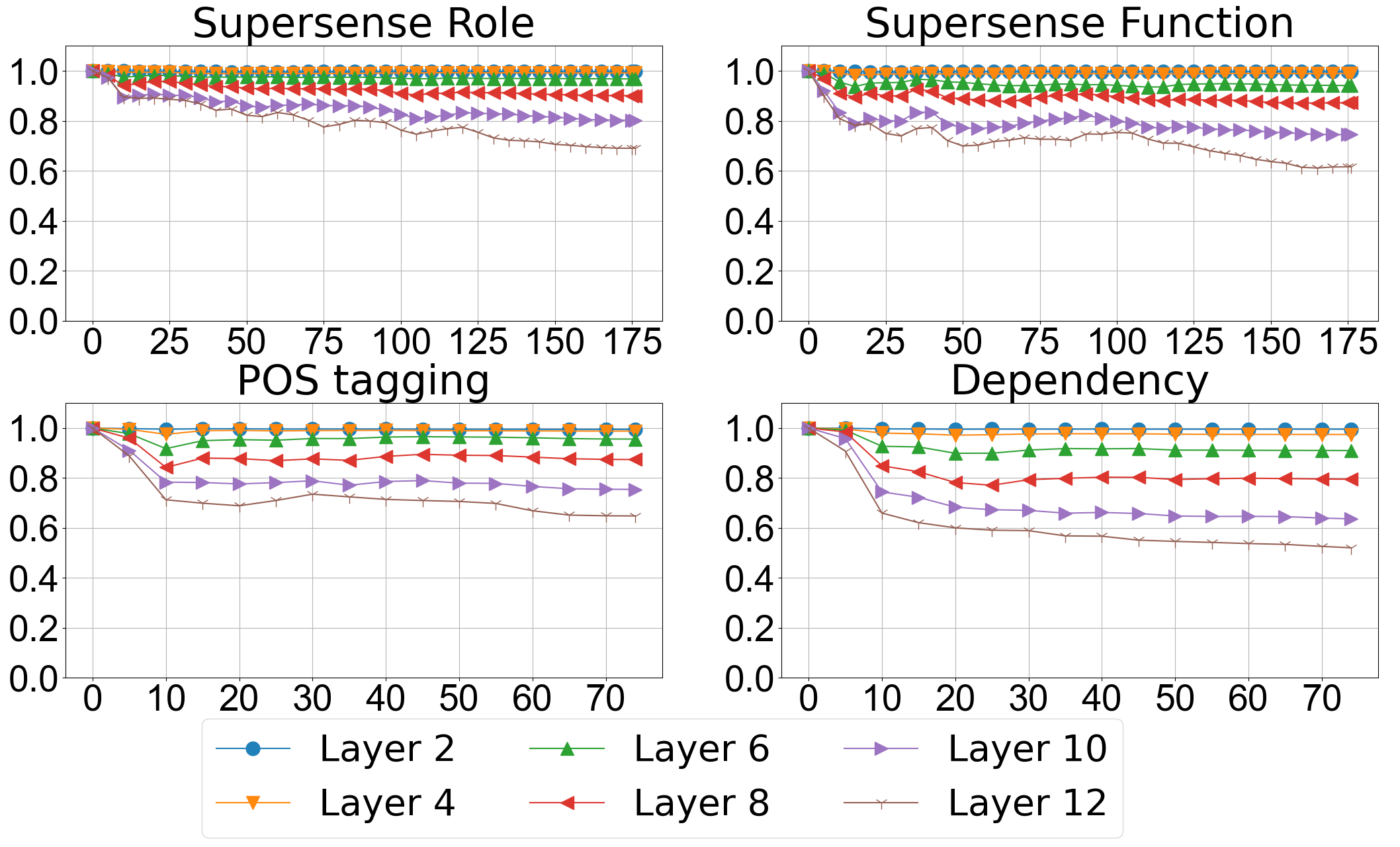}
    \caption{Dynamics of spatial similarity during the
    fine-tuning process based on \bertBase. The horizontal
    axis is the number of updates during
    fine-tuning. The vertical axis is the Pearson
    correlation coefficient between current space and its
    original version (before fine-tuning).}\label{fig:dynamic_sim}
\end{figure}

\Cref{fig:dynamic_sim} shows the results for all four
tasks.\footnote{We exclude the TREC-50 task because it is
non-linear. We cannot have the distance vectors for
non-linear representations.}  To avoid visual clutter, we
only show the plots for every alternate layer. For the
higher layers, we find that the Pearson correlation
coefficient between the original representation and the
fine-tuned one is surprisingly high (more
than $0.5$), reinforcing the notion that fine-tuning does not change
the representation arbitrarily. Instead, it attempts to  preserve the relative positions the labels.  This means the fine-tuning process encodes
task-specific information, yet it largely preserves the pre-trained
information encoded in the representation.

\begin{figure}[h]
    \centering
    \includegraphics[width=0.45\textwidth]{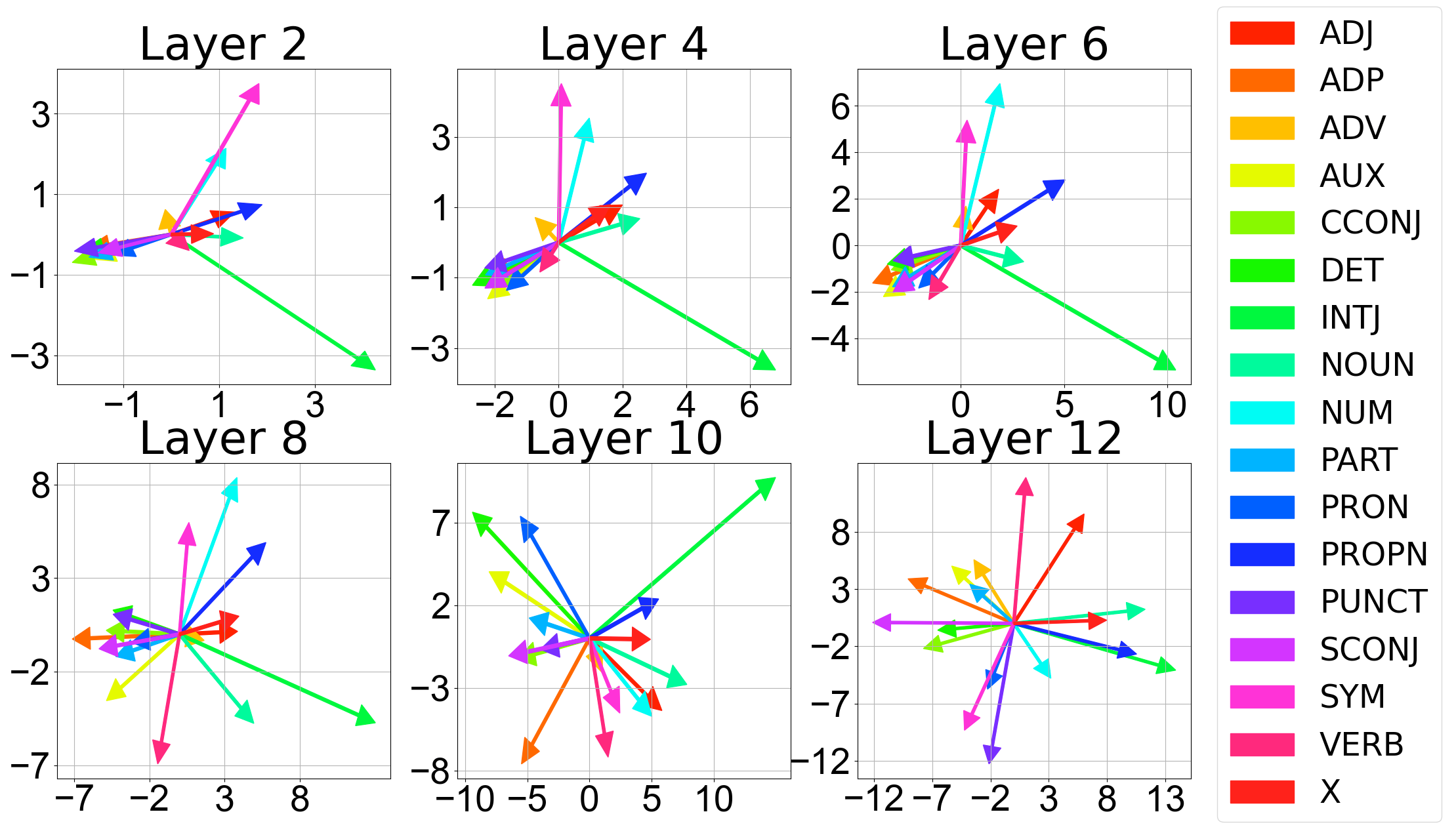}
    \caption{The PCA projection of the difference vector
    between the centroids of labels before and after
    fine-tuning based on POS tagging task and \bertBase.
    Lower layers have a much smaller projection range than
    the higher layers. This figure is best seen in color.}\label{fig:difference}
\end{figure}

\paragraph{The labels of lower layers move only in a small
region and almost in the same directions.} The unchanged
nature of lower layers raises the question: do they not
change at all? To answer this question, for every label, we
compute difference between its centroids 
before and after fine-tuning.  \Cref{fig:difference}
shows the PCA projection in 2D of these difference vectors.
For brevity, we only present the plots for every
alternative layer. A plot with all layers can be found in
\Cref{sec:app-pca-movements}.  We observe that the movements
of labels in lower layers concentrate in a few directions
compared to the higher layers, suggesting the labels in
lower layers do change, but do not separate the labels as
much as higher layers.
Also, we observe that the labels
\textsc{INTJ} and \textsc{SYM} have distinctive directions
in the lower layers.

Note that, in \Cref{fig:difference},
the motion range of lower layers is much smaller than the
higher layers.
The projected two dimensions range from $-1$
to $3$ and from $-3$ to $3$ for layer two, while for layer
12 they range from $-12$ to $13$ and $-12$ to $8$,
suggesting that labels in lower layers only move in a small
region compared to higher layers.
\Cref{fig:pca_paths} shows an example of this difference.
Compared to the layer 12 (right) paths, we see that the
layer 1 paths (left) traverse almost the same trajectories,
which is consistent with the observations from \Cref{fig:difference}.




\section{Discussion}\label{sec:discussion}

\noindent\textbf{Does fine-tuning always improve
performance?} Indeed, fine-tuning almost always improves
task performance. However, rare cases exist where
fine-tuning decreases the performance.  Fine-tuning
introduces a divergence between the training set and unseen
examples (\Cref{sec:fine-tuned-performance}).  However, it
is unclear how this divergence affects the generalization
ability of representations, \eg does this divergence suggest
a new kind of overfitting that is driven by representations
rather than classifiers? 

\noindent\textbf{How does fine-tuning alter the
representation to adjust for downstream tasks?} Fine-tuning
alters the representation by grouping points with the same
label into small number of clusters (\Cref{sec:linear}) and pushing each
label cluster away from the others (\Cref{sec:findings-strucure}). We hypothesize that the distances
between label clusters correlate with the classification
performance and confirm this hypothesis by investigating
cross-task fine-tuning (\Cref{sec:cross-tasks}). Our
findings are surprising because fine-tuning for a
classification task does not need to alter the
geometry of a representation if the data is already linearly separable in the
original representation. 
What we observe
reveals geometric
properties that characterize good representations.
We do not show theoretical analysis to connect our geometric
findings to representation learnability, but the findings in
this work may serve as a starting point for a learning theory
for representations.

\noindent\textbf{How does fine-tuning change the underlying
geometric structure of different layers?} It is established
that higher layers change more than the lower ones. In this
work, we analyze this behavior more closely.  We discover that
higher layers do not change arbitrarily; instead, they
remain similar to the untuned version. Informally, we can
say that fine-tuning only ``slightly'' changes even the
higher layers (\Cref{sec:layers}). Nevertheless, our analysis
does not reveal why higher layers change more than the lower
layers.  A deeper analysis of model parameters during
fine-tuning is needed to understand the difference between
lower and higher layers.

\noindent\textbf{Limitations of this work.} Our experiments
use the BERT family of models for English tasks. Given the architectural
similarity of transformer language models, we may be able to extrapolate the
results to other models, but further work is needed to confirm our findings to
other languages or model architectures. 
In our analysis, we
ignore the structure within each cluster, which is another information source
for studying the representation. We plan to investigate these aspects in future
work. We make our code available for replication and extension by the community.


\section{Related Work}\label{sec:related}

There are many lines of work that focus on
analyzing and understanding representations. The most
commonly used technique is the classifier-based method.
Early
work~\citep{DBLP:conf/iclr/AlainB17,kulmizev-etal-2020-neural}
starts with using linear classifiers as the probe. \citet{hewitt-liang-2019-designing} pointed
out that a linear probe is not sufficient to evaluate a
representation. Some recent work also employ non-linear
probes~\citep{DBLP:conf/iclr/TenneyXCWPMKDBD19,
eger-etal-2019-pitfalls}.
There are also efforts to inspect the representations from a
geometric
persepctive~\citep[\eg][]{ethayarajh-2019-contextual,mimno-thompson-2017-strange},
including the recently proposed
\directprobe~\citep{zhou-srikumar-2021-directprobe}, which we use in
this work. Another line of probing work  designs control
tasks~\citep{ravichander-etal-2021-probing,DBLP:conf/iclr/LanCGGSS20}
to reverse-engineer the internal mechanisms of
representations~\citep{kovaleva-etal-2019-revealing,wu-etal-2020-perturbed}.
However, in contrast to our work, most
studies~\cite{zhong-etal-2021-factual,li-etal-2021-bert,DBLP:conf/iclr/ChenFXXTCJ21} focused on
 pre-trained representations, not fine-tuned ones.

While fine-tuning pre-trained representations usually
provides strong empirical
performance~\citep{wang-etal-2018-glue,DBLP:journals/tacl/TalmorEGB20},
how fine-tuning manage to do so has
remained an open question. Moreover, the
instability~\citep{DBLP:journals/corr/abs-2006-04884,DBLP:journals/corr/abs-2002-06305,zhang2020revisiting}
and forgetting
problems~\citep{chen-etal-2020-recall,he-etal-2021-analyzing}
make it harder to analyze fine-tuned
representations. Despite these difficulties, previous
work~\citep{merchant-etal-2020-happens,mosbach-etal-2020-interplay,hao-etal-2020-investigating}
draw valuable conclusions about fine-tuning. This work
extends this line of effort and provides a deeper
understanding of how fine-tuning changes
representations.


\section{Conclusions}\label{sec:conclusion} In this work, we
take a close look at how fine-tuning a contextualized representation for a task modifies it.
We investigate the fine-tuned
representations of several BERT models using two probing
techniques: classifier-based probing and \directprobe.
First, we show that fine-tuning introduces divergence between training and test
set, and in at least one case, hurts generalization.  Next, we show fine-tuning
alters the geometry of a representation by pushing points belonging to the same
label closer to each other, thus simpler and better classifiers.  We confirm
this hypothesis by cross-task fine-tuning experiments. Finally, we discover that
while adjusting representations to downstream tasks, fine-tuning largely
preserves the original spatial structure of points across all layers.
Taken collectively, the empirical study presented in this work can
not only justify the impressive performance of fine-tuning, but may also lead to a better understanding of learned representations.


\section*{Acknowledgments}
\label{sec:acknowledgments}

We thank the ARR reviewers and the Utah NLP group for their constructive
feedback. This work is partially supported by NSF grants \#1801446 (SaTC) and
\#1822877 (Cyberlearning), and a generous gift from Verisk Inc.


\bibliography{anthology,custom}
\bibliographystyle{acl_natbib}

\appendix

\section{Fine-tuning Details}\label{sec:fine-tuning-details}
In this work, we fine-tune all tasks and representations
using HuggingFace library. We use a linear weight schduler
with a learning rate of $3e^{-4}$, which uses $10\%$ of the
total update steps as the warmup steps. The same schduler is
used for all tasks. All the models are optimized by
Adam~\cite{kingma2014adam} with batch size of 32. All the
fine-tuning is run on a single Titan GPU. The best hidden-layer sizes for each
task are shown in~\Cref{tb:complete}.

\section{Summary of Tasks}\label{sec:task-summarization}
In this work, we conduct experiments on five NLP tasks, which
are chosen to cover different usages of the representations we study.
Table~\ref{tb:tasks} summarizes these tasks.

\begin{table*}[]
\small
\centering
\begin{tabular}{@{}lrrrccccc@{}}
\toprule
    Task & \#Training & \#Test &\#Labels  & Token-based & Sentence-based & Pair-wise & Semantic & Syntax \\ \midrule
    Supersense-role & \num{4282} & \num{457} &\num{47} & $\surd$ & & & $\surd$ & \\
    Supersense-function & \num{4282} & \num{457} & \num{40}
    & $\surd$ & & & $\surd$ & \\
POS & \num{16860} & \num{4323} & \num{17} & $\surd$ & & & & $\surd$ \\
    Dependency Relation  & \num{16054} & \num{4122} &
    \num{46}& & & $\surd$ & & $\surd$ \\ 
TREC-50  & \num{5452} & \num{500}& \num{50} & & $\surd$ &  & $\surd$ & \\ \bottomrule
\end{tabular}
\caption{Statistics of the five tasks with their different
    characteristics.}\label{tb:tasks}
\end{table*}

\section{Probing Performance}\label{sec:app-probing-perf}

\Cref{tb:complete} shows the complete table of probing
results in our experiments. The last column is the spatial
similarity between the training set and test set. Some entries
are missing because the similarity can only be computed
on the representations that are linearly separable for the
given task.


\begin{table*}[]
\small
\centering
\begin{tabular}{@{}lllrrrrrr@{}}
\toprule
Representations              & Task                     &
    & Acc   & Std & Best Layer Size  & \#Cluster & is Linear & Similarity \\ \midrule
\multirow{8}{*}{\bertTiny}   & \multirow{2}{*}{POS}     & original   & 90.76 & 0.24 & (256, 64) & 3936      & N         & -              \\
                             &                          & fine-tuned & 91.67 & 0.29 & (64, 64) &  20        & N         & -              \\ \cmidrule{2-9}
                             & \multirow{2}{*}{DEP}     & original   & 86.74 & 0.22 & (256, 256) & 653       & N         & -              \\
                             &                          & fine-tuned & 89.04 & 0.20 & (256, 256) & 46        & Y         & 0.88           \\  \cmidrule{2-9}
                             & \multirow{2}{*}{PS-fxn} &  original    & 74.14 & 1.42 & (256, 256) & 402       & N         & -              \\
                             &                          & fine-tuned & 74.40 & 0.68 & (256, 128) & 40        & Y         & 0.72           \\  \cmidrule{2-9}
                             & \multirow{2}{*}{PS-role} & original   & 58.38 & 0.78 & (256, 64) &  46        & Y         & 0.76           \\
                             &                          & fine-tuned & 60.31 & 0.29 & (64, 64) & 46        & Y         & 0.70           \\  \cmidrule{2-9}
                             & \multirow{2}{*}{TREC-50} & original   & 68.12 & 0.82 & (256, 256)  & 399       & N         & -           \\
                             &                          & fine-tuned & 84.04 & 0.93 & (256, 256) & 51        & N         & -           \\     \midrule
\multirow{8}{*}{\bertMini}   & \multirow{2}{*}{POS}     & original   & 93.81 & 0.10  & (256, 32) & 2429      & N         & -              \\
                             &                          & fine-tuned & 94.91 & 0.03 & (256, 32) & 17        & Y         & 0.70           \\ \cmidrule{2-9}
                             & \multirow{2}{*}{DEP}     & original   & 91.82 & 0.09 & (256, 128) & 46        & Y         & 0.93           \\
                             &                          & fine-tuned & 93.55 & 0.07 & (256, 128) & 46        & Y         & 0.86           \\  \cmidrule{2-9}
                             & \multirow{2}{*}{PS-fxn}  & original   & 82.45 & 1.07 & (256, 256) & 40        & Y         & 0.77           \\
                             &                          & fine-tuned & 84.25 & 0.39 & (256, 128) & 40        & Y         & 0.53           \\  \cmidrule{2-9}
                             & \multirow{2}{*}{PS-role} & original   & 68.05 & 1.08 & (256, 256) & 46        & Y         & 0.81           \\
                             &                          & fine-tuned & 71.90 & 1.06 & (256, 64) & 46        & Y         & 0.59           \\ \cmidrule{2-9}
                             & \multirow{2}{*}{TREC-50} & original   & 74.12 & 1.25 & (256, 256) & 127       & N         & -              \\
                             &                          & fine-tuned & 88.36 & 0.50 & (64, 32) & 52        & N         & -              \\ \midrule
\multirow{8}{*}{\bertSmall}  & \multirow{2}{*}{POS}     & original   & 94.26 & 0.13 & (256, 32) & 17        & Y         & 0.96           \\
                             &                          & fine-tuned & 95.43 & 0.06 & (128, 64) & 17        & Y         & 0.72           \\  \cmidrule{2-9}
                             & \multirow{2}{*}{DEP}     & original   & 92.93 & 0.14 & (256, 64) & 46        & Y         & 0.93           \\
                             &                          & fine-tuned & 94.48 & 0.14 & (256, 64) & 46        & Y         & 0.78           \\  \cmidrule{2-9}
                             & \multirow{2}{*}{PS-fxn}  & original   & 86.26 & 0.54 & (256, 256) & 40        & Y         & 0.82           \\
                             &                          & fine-tuned & 85.08 & 0.35 & (256, 256) & 40        & Y         & 0.44           \\  \cmidrule{2-9}
                             & \multirow{2}{*}{PS-role} & original   & 74.22 & 1.03 & (256, 256) & 46        & Y         & 0.84           \\
                             &                          & fine-tuned & 74.57 & 0.61 & (128, 128) & 46        & Y         & 0.54           \\  \cmidrule{2-9}
                            & \multirow{2}{*}{TREC-50} & original   & 81.32 & 0.61 & (256, 128) &  113       & N         & -           \\
                             &                         & fine-tuned & 89.60 & 0.22 & (256, 64) &  51        & N         & -              \\  \midrule
\multirow{8}{*}{\bertMedium} & \multirow{2}{*}{POS}     & original   & 94.40 & 0.08 & (256, 128) &  17        & Y         & 0.97           \\
                             &                          & fine-tuned & 95.56 & 0.05 & (64, 32)  &  17        & Y         & 0.67           \\ \cmidrule{2-9}
                             & \multirow{2}{*}{DEP}     & original   & 92.54 & 0.14 & (256, 256) &  46        & Y         & 0.94           \\
                             &                          & fine-tuned & 94.76 & 0.20 & (128, 128) & 46        & Y         & 0.79           \\  \cmidrule{2-9}
                             & \multirow{2}{*}{PS-fxn}  & original   & 86.56 & 0.41 & (256, 128) & 40        & Y         & 0.80           \\
                             &                          & fine-tuned & 88.45 & 0.45 & (128, 256) & 40        & Y         & 0.59           \\  \cmidrule{2-9}
                             & \multirow{2}{*}{PS-role} & original   & 76.28 & 1.00 & (256, 32) & 46        & Y         & 0.83           \\
                             &                          & fine-tuned & 78.86 & 0.58 & (128, 128) & 46        & Y         & 0.58           \\   \cmidrule{2-9}
                             & \multirow{2}{*}{TREC-50} & original   & 80.68 & 1.16& (256, 64) & 110       & N         & -              \\
                             &                          & fine-tuned & 89.80 & 0.33 & (32, 64) & 52        & N         & -              \\  \midrule
\multirow{8}{*}{\bertBase}   & \multirow{2}{*}{POS}     & original   & 93.39 & 0.31 & (256, 128) & 17        & Y         & 0.97           \\
                             &                          & fine-tuned & 95.68 & 0.02 & (128, 64) & 17        & Y         & 0.70           \\ \cmidrule{2-9}
                             & \multirow{2}{*}{DEP}     & original   & 89.39 & 0.08 & (256, 128) & 46        & Y         & 0.92           \\
                             &                          & fine-tuned & 94.76 & 0.05 & (64, 256) & 46        & Y         & 0.76           \\  \cmidrule{2-9}
                             & \multirow{2}{*}{PS-fxn}  & original   & 87.75 & 0.41 & (256, 128) & 40        & Y         & 0.84           \\
                             &                          & fine-tuned & 89.58 & 0.67 & (32, 256) & 40        & Y         & 0.57           \\  \cmidrule{2-9}
                             & \multirow{2}{*}{PS-role} & original   & 74.49 & 0.84 & (256, 128) & 46        & Y         & 0.82           \\
                             &                          & fine-tuned & 81.14 & 0.26 & (256, 128) & 46        & Y         & 0.52           \\ \cmidrule{2-9}
                             & \multirow{2}{*}{TREC-50} & original   & 85.24 & 0.85 & (256, 128) & 162       & N         & -              \\
                             &                          & fine-tuned & 90.36 & 0.32 & (64, 32) & 51        & N         & -              \\ \bottomrule

\end{tabular}\caption{A complete table of the probing
    results of five representations on five
    tasks.}\label{tb:complete}
\end{table*}

\section{Dynamics of Minimum Distances}\label{sec:app-dynamic-min-dis}

\begin{figure*}
    \centering
    \includegraphics[width=1.0\textwidth]{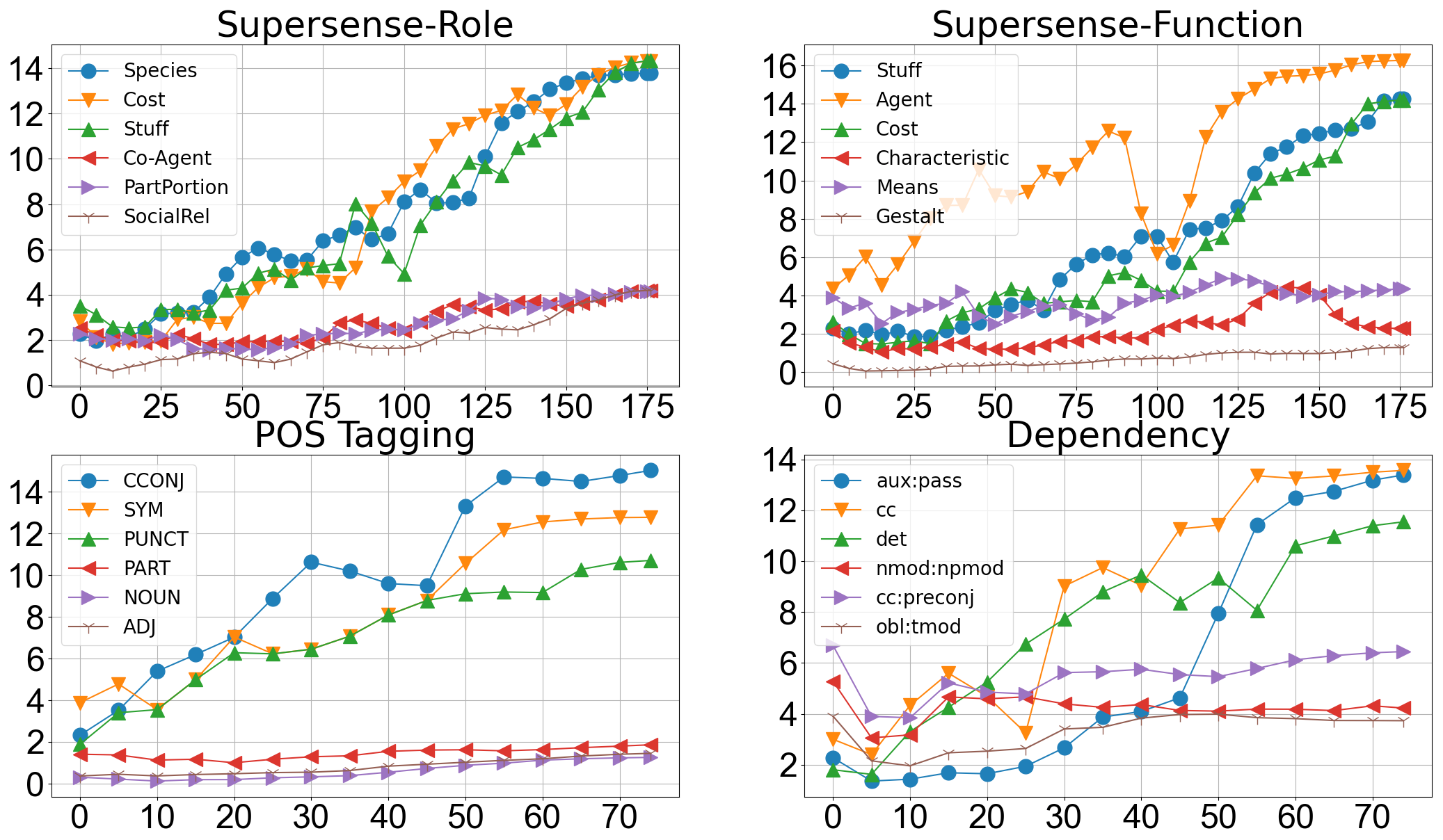}
    \caption{The dynamics of the minimum distance of the
    three labels where the distance increases the most, and
    three labels where is increases the least.
    The horizontal axis is the number of fine-tuning
    updates; the vertical axis is chosen label's minimum
    distance to other labels. These results come
    from the last layer of
    \bertBase.}\label{fig:ss-all-movements}
\end{figure*}

\Cref{fig:ss-all-movements} shows the dynamics of minimum
distances for labels on all four tasks. For clarity, we only
present the distances for the three labels where the
distances increase the most and the three where it decreases
the most. 

\section{PCA Projections of the Movements}\label{sec:app-pca-movements}
\begin{figure*}
    \centering
    \includegraphics[width=1.0\textwidth]{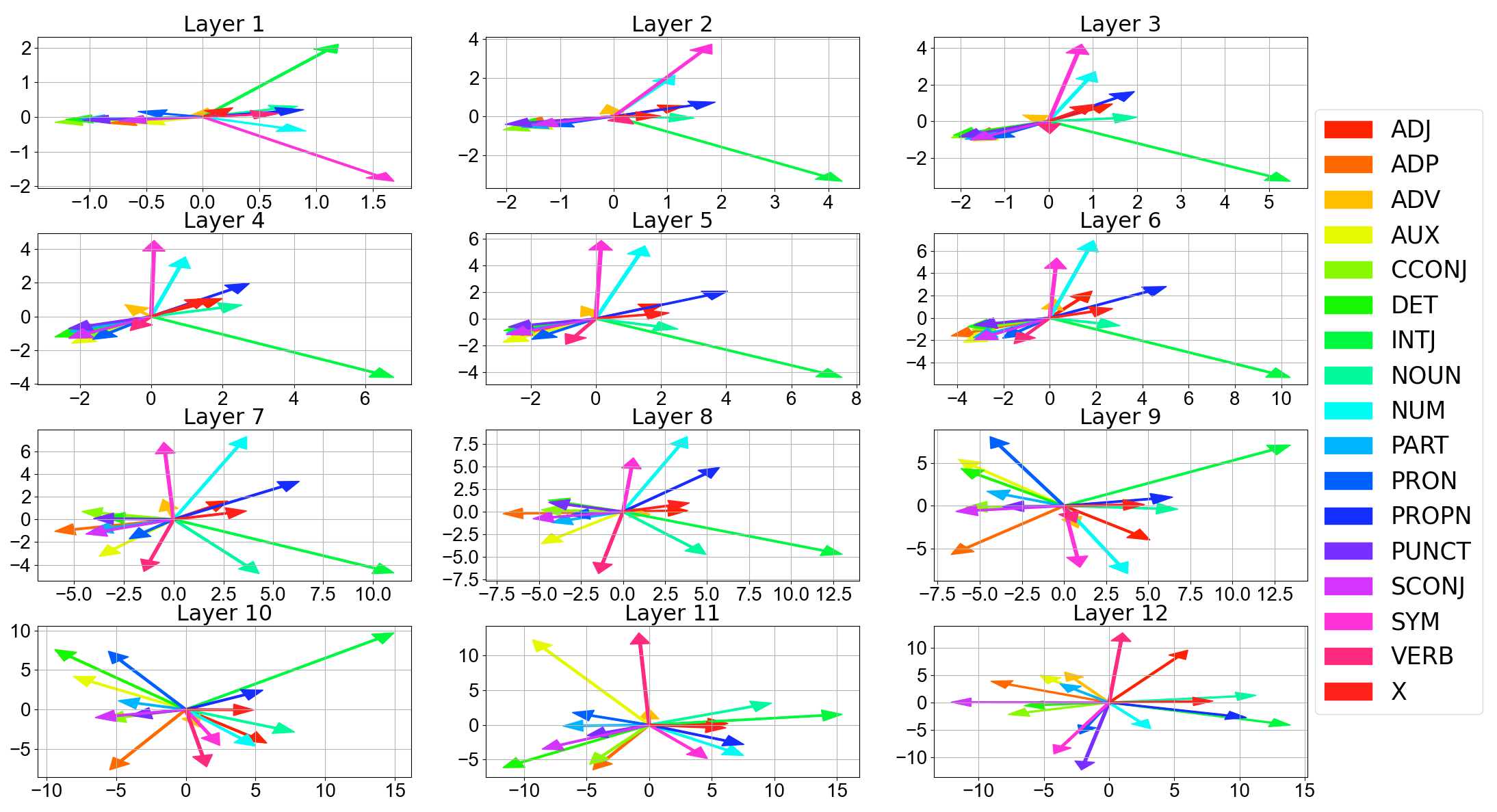}
    \caption{The PCA projection of the difference vector
    between the centroids of labels before and after
    fine-tuning based on POS tagging task and \bertBase.}\label{fig:pos-pud-diff}
\end{figure*}

\begin{figure*}
    \centering
    \includegraphics[width=1.0\textwidth]{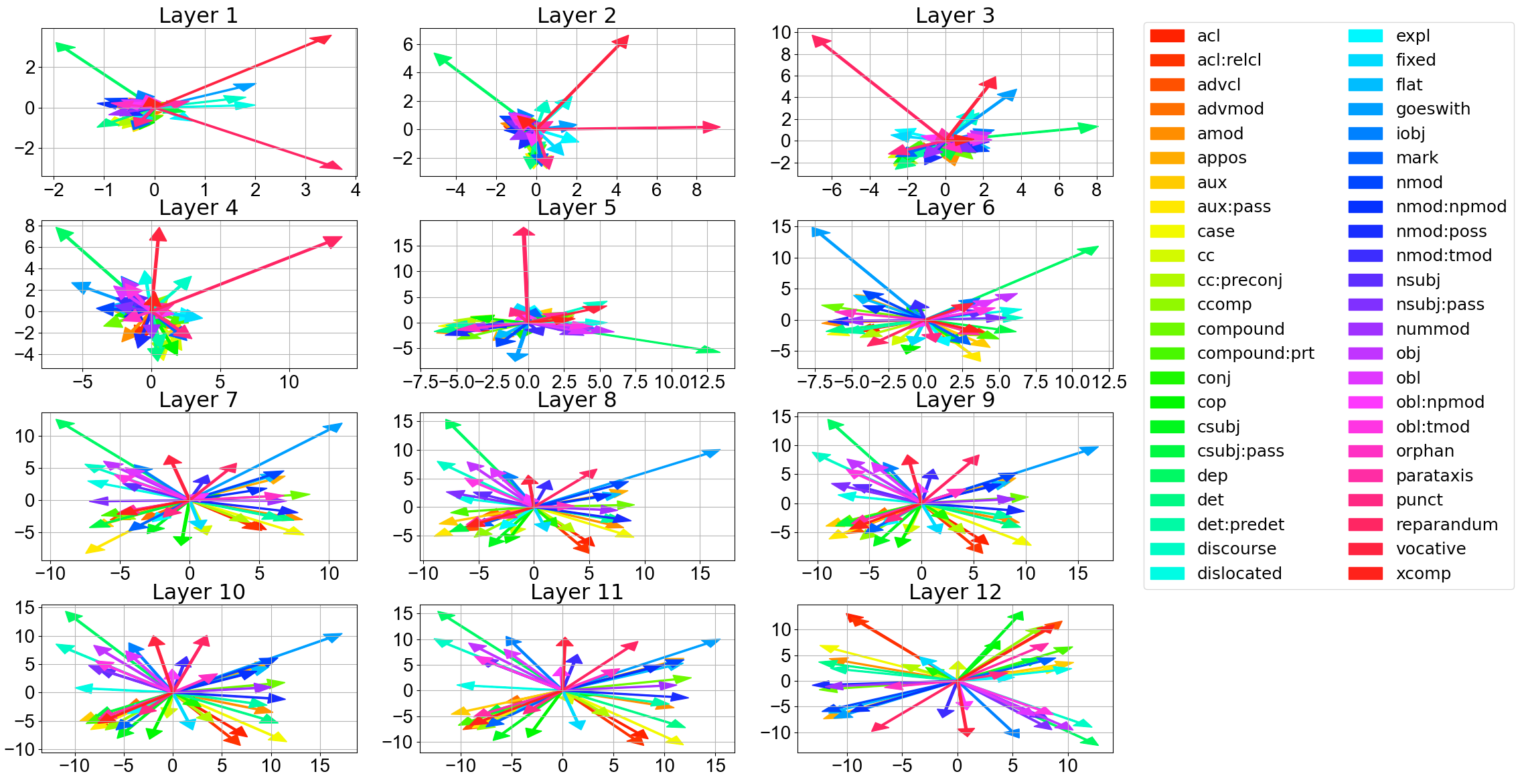}
    \caption{The PCA projection of the difference vector
    between the centroids of labels before and after
    fine-tuning based on dependency prediction task and \bertBase.}\label{fig:dep-pud-diff}
\end{figure*}

\begin{figure*}
    \centering
    \includegraphics[width=1.0\textwidth]{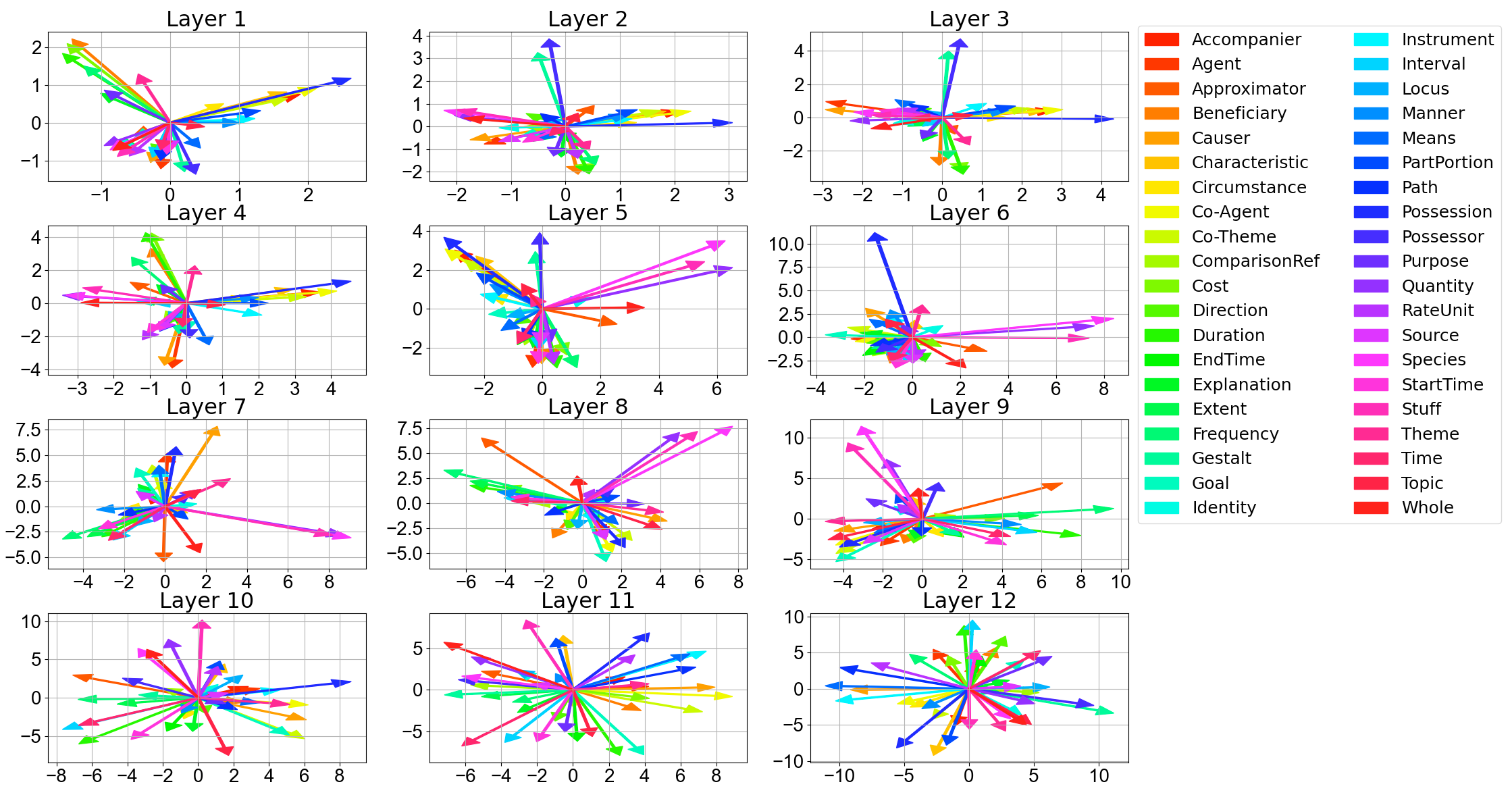}
    \caption{The PCA projection of the difference vector
    between the centroids of labels before and after
    fine-tuning based on Supersense function task and \bertBase.}\label{fig:ss-func-diff}
\end{figure*}

\begin{figure*}
    \centering
    \includegraphics[width=1.0\textwidth]{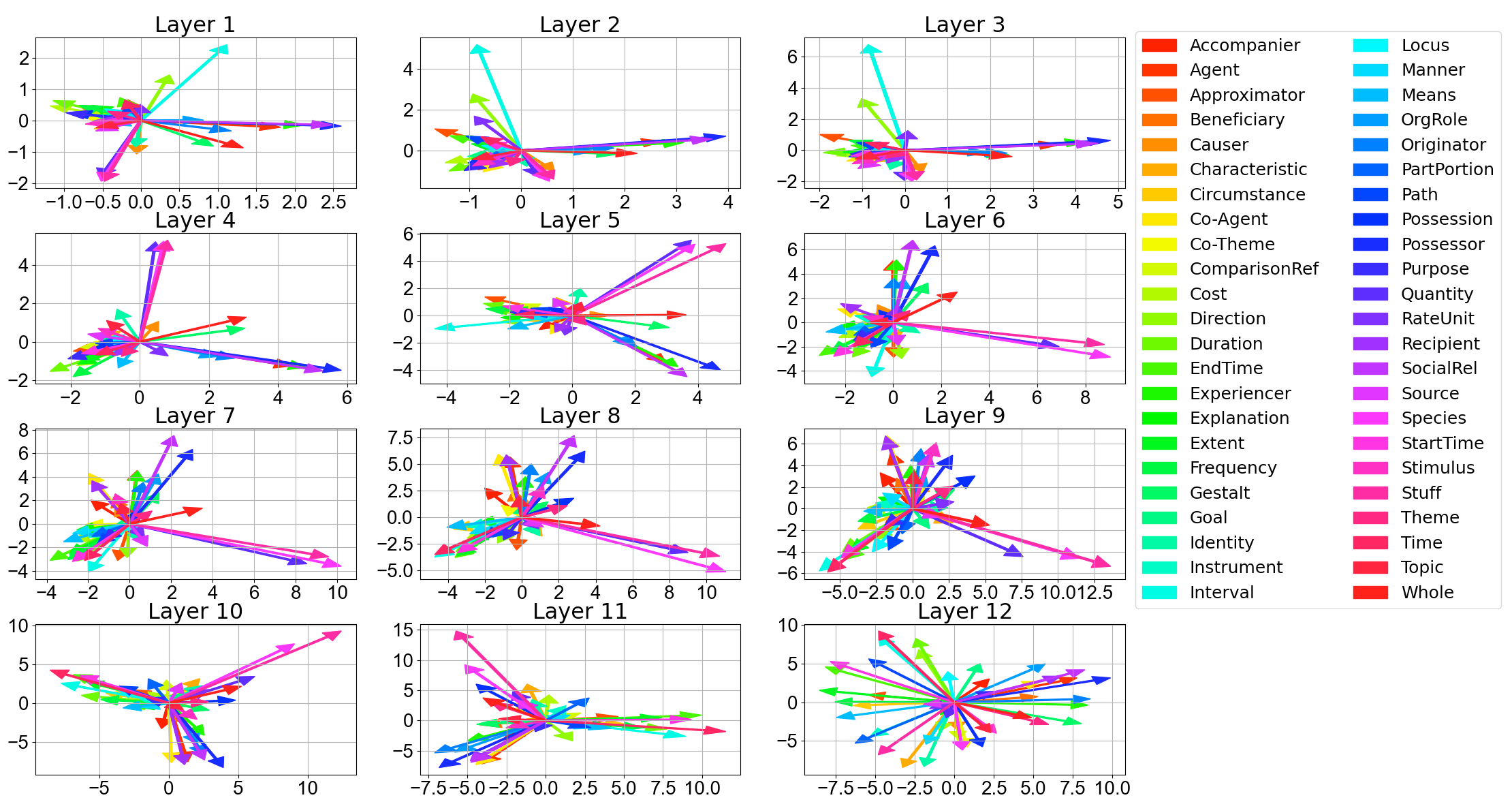}
    \caption{The PCA projection of the difference vector
    between the centroids of labels before and after
    fine-tuning based on Supersense role task and \bertBase.
    }\label{fig:ss-role-diff}
\end{figure*}

Figures~\ref{fig:pos-pud-diff}--\ref{fig:ss-role-diff} show the
PCA projections of the difference vector between the
centroids of labels before and after fine-tuning based on
\bertBase.

\end{document}